\documentclass[letterpaper, 10 pt, conference]{ieeeconf}  
\IEEEoverridecommandlockouts                              

\usepackage{float}

\usepackage{amsmath} 
\usepackage{amssymb}  
\usepackage{graphics}
\usepackage{graphicx}
\usepackage{multirow}

\usepackage{xparse}
\NewDocumentCommand{\ceil}{s O{} m}{%
  \IfBooleanTF{#1} 
    {$\left\lceil#3\right\rceil$} 
    {#2\lceil#3#2\rceil} 
}

\usepackage{todonotes}
\usepackage{amsmath}
\DeclareMathOperator*{\argmax}{arg\,max}

\usepackage{amsmath,amssymb}
\DeclareMathOperator{\E}{\mathbb{E}}

\usepackage{bbm}

\usepackage{algorithm,algorithmicx}
\usepackage{algpseudocode}


\usepackage{algorithm,algorithmicx}
\usepackage{algorithmicx}
\usepackage{algorithm} 
\usepackage{algpseudocode}
\makeatletter
\let\NAT@parse\undefined
\makeatother
\usepackage[numbers,sort&compress]{natbib}

\title{\LARGE \bf
Learning to Scaffold the Development of Robotic Manipulation Skills}

\author{Lin Shao, Toki Migimatsu and Jeannette Bohg
\thanks{All authors are with the Stanford Artificial Intelligence Lab (SAIL), Stanford University, CA, USA. {\tt\footnotesize [lins2,takatoki,bohg]
@stanford.edu}}%
\thanks{This work has been partially supported by JD.com American Technologies Corporation (“JD”) under the SAIL-JD AI Research Initiative. This article solely reflects the opinions and conclusions of its authors and not JD or any entity associated
 with JD.com. Toyota Research Institute (”TRI”) provided
funds to assist the authors with their research but this article solely reflects
the opinions and conclusions of its authors and not TRI or any other Toyota entity.}}

\begin{document}

\maketitle
\thispagestyle{empty}
\pagestyle{empty}

\begin{abstract}
Learning contact-rich, robotic manipulation skills is a challenging problem due to the high-dimensionality of the state and action space as well as uncertainty from noisy sensors and inaccurate motor control. To combat these factors and achieve more robust manipulation, humans actively exploit contact constraints in the environment. By adopting a similar strategy, robots can also achieve more robust manipulation. In this paper, we enable a robot to autonomously modify its environment and thereby discover how to ease manipulation skill learning. Specifically, we provide the robot with fixtures that it can freely place within the environment. These fixtures provide hard constraints that limit the outcome of robot actions. Thereby, they funnel uncertainty from perception and motor control and {\em scaffold\/} manipulation skill learning. We propose a learning system that consists of two learning loops. In the outer loop, the robot positions the fixture in the workspace. In the inner loop, the robot learns a manipulation skill and after a fixed number of episodes, returns the reward to the outer loop. Thereby, the robot is incentivised to place the fixture such that the inner loop quickly achieves a high reward. We demonstrate our framework both in simulation and in the real world on three tasks: peg insertion, wrench manipulation and shallow-depth insertion. We show that manipulation skill learning is dramatically sped up through this way of scaffolding.  \label{sec:abstract}
\end{abstract}

\section{Introduction}
A hallmark in robotics research is the autonomous learning of skills. If achieved, this ability would make the deployment of robots more flexible by alleviating the need to adapt the environment to match the limited abilities of the robot. The challenges towards this goal are the high dimensionality of the state and action space as well as uncertainty from perception and motor control. \citet{deimel2016exploitation} have shown that humans exploit contact constraints in the environment to combat these effects. The environment provides physical constraints on an action and funnels uncertainty due to noise in perception and control. There also has been work that shows how this strategy makes robotic manipulation more robust~\citep{deimel2016exploitation,kazemi2012robust,righetti2014autonomous,hudson2012end,toussaint2014dual,chavan2015prehensile}. For example, fixtures is a widely used practice in industry for various application such as machining, assembly and inspection. The principles and designs of fixtures have been extensively studied~\cite{asada1985kinematic,chou1989mathematical}. 

However, these works typically consider only a limited set of constraints and predefine how they can be exploited. In this paper, we enable a robot to actively modify its environment and discover different ways in which constraints can be exploited. Specifically, we provide the robot with fixtures that it can freely place in the workspace to act as physical scaffolding for manipulation skill learning. In this way, our learning framework mirrors educational scaffolding used in human learning. The proposed approach consists of two learning loops: an outer loop places a fixture in the workspace, and then an inner loop attempts to learn a manipulation skill through reinforcement learning. The reward achieved by the inner loop is returned to the outer loop so that the outer loop can optimise the fixture placement for the next trial. In this way, the outer loop has an incentive to optimise the fixture such that the inner loop achieves a high reward quickly.

\begin{figure}[tb!]
 \centering
 \includegraphics[width=1.0\linewidth ]{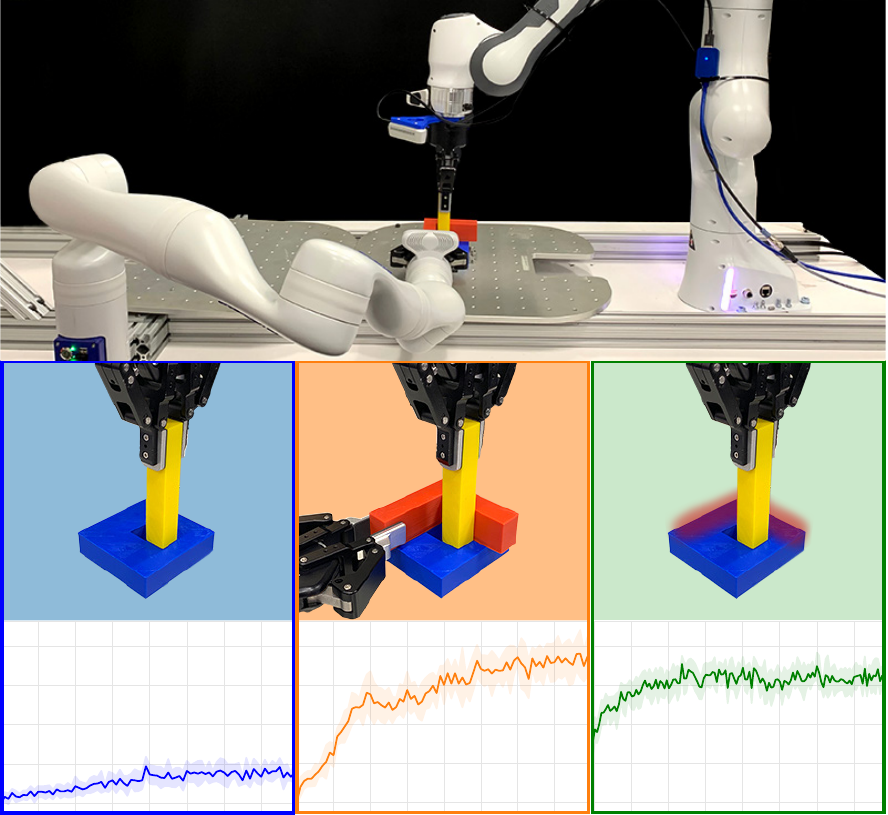}
 \caption{
Our proposed learning framework enables robots to autonomously use fixtures to aid with manipulation skill learning. The top image shows our experimental setup with one manipulator arm to position the fixture and another to complete the manipulation task. The bottom images compare the learning curves of the manipulation task without any fixture (left), with a fixture (middle), and with a virtual potential field to replace the fixture as the fixture is gradually moved away (right). The fixture dramatically improves the robot's ability to learn the task.
}
\label{fig:teaser}
\end{figure}


Our primary contributions are: 1) We propose a learning framework for robots to leverage fixtures to assist with complex manipulation tasks. 2) We introduce an algorithm that improves sample efficiency in bandit problems with continuous action spaces and discontinuous reward functions. 3) We demonstrate that our method dramatically improves robot performance on three challenging manipulation tasks (peg insertion, wrench manipulation, and shallow-depth insertion~\cite{kim2019shallow}) both in simulation and the real world. 4) We propose a method that allows the robot to maintain task performance while gradually removing the fixtures from the environment.

\label{sec:intro}

\section{Related Work}
\subsection{Learning to Leverage the Environment}
Exploiting contacts and environmental constraints to reduce uncertainty during manipulation has received many researchers' attentions. \citet{kazemi2012robust} show that contact with support surfaces is critical for grasping small objects and design a closed-loop hybrid controller that mimics the pre-grasp and landing strategy for finger placement. ~\citet{deimel2016exploitation} demonstrate how humans produce robust grasps by exploiting constraints in the environment.  ~\citet{righetti2014autonomous} propose an architecture to utilize contact interactions for force/torque control and optimization-based motion planning in grasping and manipulation tasks. ~\citet{hudson2012end} present a model-based approach to improve system knowledge about the combined robot and environmental state through deliberate interactions with the objects and the environment. ~\citet{toussaint2014dual} integrate the idea of exploiting contacts with trajectory optimizations.

In all the above works, robots interact with the environment using a limited set of constraints and a predefined way to exploit them. In contrast, we propose a data-driven learning framework for robots to automatically discover and learn how to change and leverage the environment with self-supervision. Our framework not only allows robots to exploit environmental constraints passively but also to actively create new constraints using fixtures to generate "funnels" for optimal learning convergence. After learning how to perform the manipulation tasks with fixtures, our framework is also able to learn how to maintain task performance with the fixtures removes.

\subsection{Learning Robotic Manipulation Skills}
Our framework can be adopted to improve the performance of a range of manipulation tasks including robotic assembly, tool manipulation, in-hand manipulation, and re-grasping. We review the related work in these domains.
\subsubsection{Insertion} \citet{inoue2017deep} separate the peg-in-hole process into two stages, searching and inserting, and use LSTMs to learn separate policies for each one.
\citet{thomas2018learning} combine reinforcement learning with a motion planner to shape the state cost in a high-precision setting. 
\citet{luo2019reinforcement} utilize deep reinforcement learning to learn variable impedance controllers for assembly tasks. \citet{lee2019making} combine both vision and touch to learn a policy for insertion.

\subsubsection{Tool Manipulation}
\citet{zhu2015understanding} consider handheld physical tools like hammers and utilize RGB-D images to identify functional and affordance areas of the tools. \citet{fang2018learning} learn task-oriented grasps for tools. 
In our work, we assume that the tool is already grasped and focus on discovering how to exploit fixtures for scaffolding the skill learning process.

\subsubsection{In-hand Manipulation and Re-Grasping} \citet{chavan2015prehensile} explore the manipulation of a grasped object by pushing it against the environment. 
\citet{kim2019shallow} use in-hand manipulation to insert a flat object into a hole with a shallow depth e.g. a battery into a mobile phone. 

Unlike all the approaches mentioned above, we focus on actively changing the environment to support manipulation skill learning. These approaches can be placed in the inner loop of our learning framework to improve the performance of each individual manipulation task. Our outer loop would then learn how to use fixtures to improve the performance of the inner loop policies.

\subsection{Contextual Bandits with Continuous Action and Discontinuous Reward}
We formulate the problem of optimally placing the fixture as a contextual bandits problem with a continuous action space and discontinuous reward. The aim is to maximize the cumulative rewards over a series of trials. For these types of bandit problems, prior work typically assumes that nearby actions have similar rewards. For example, the rewards are assumed to be Lipschitz continuous as a function of the actions.
\citet{kleinberg2013bandits} propose the zooming algorithm to adaptively discretize and sample within the continuous action space. However, in our case and in many other real-world applications, the function which maps actions to rewards is discontinuous. In the peg-in-hole task, for example, if the fixture is precisely aligned with the boundary of the hole as shown in Fig.~\ref{fig:Qmap}, then the robot can successfully insert the peg into the hole and achieve a high reward. However, if the fixture is slightly moved such that it blocks the hole, the robot suddenly receives zero reward. 
 \citet{krishnamurthy2019contextual} address this challenge of discontinuity in the action-to-reward function by mapping each discrete action into a well-behaved distribution with no continuity assumptions. However, their proposed method, which is based on policy elimination~\cite{dudik2011efficient}, is not suitable for high dimensional spaces like images. Inspired by the ideas of adaptive zooming~\cite{kleinberg2013bandits} and smoothing~\cite{krishnamurthy2019contextual}, we present a novel \emph{Smoothed Zooming} algorithm to solve high dimensional contextual bandits with continuous actions and a discontinuous reward.

\label{sec:relatedwork}

\section{Problem Definition}
In this work, we want to enable a robot to autonomously alter its environment to ease manipulation skill learning. This requires the robot to find an optimal placement of a fixture and to learn a manipulation policy that exploits this fixture. 

Given an observation of a manipulation scene, e.g. an image denoted as $s \in \mathcal{I}$, the robot has to propose an optimal fixture pose $f^* \in \mathcal{F}$. Given $f^*$, the robot then learns a policy $\pi^* \in \Pi$ that maximizes the expected discounted rewards $\sum_{t=1}^T\gamma^{t-1}R_t$ where $\gamma$ is the discounted factor. This can be written as:
\begin{equation}
  (f^*,\pi^*) = \argmax_{f\in\mathcal{F}}\argmax_{\pi \in \Pi} \E\left[\sum_{t=1}^{T}\gamma^{t-1}R_t(s_t,\pi(s_t|f))\right]
\end{equation}

In this work, we also consider the problem of continuing to update the policy while the fixture is gradually removed. This could be considered as a physical curriculum for learning manipulation skills, similar to training wheels for learning how to bike. Formally, the robot has to learn a second optimal policy $\pi^{**} \in \Pi$ without the fixture but starting from the previous policy $\pi^{*}$:
\begin{equation}
    \pi^{**} = \argmax_{\pi \in \Pi} \E\left[\sum_{t=1}^{T}\gamma^{t-1}R_t(s_t,\pi(s_t))\right]
\end{equation}
\label{sec:relatedwork}

\section{Technical Approach}
We propose a learning system that consists of two learning loops. We use {\em Reinforcement Learning} (RL) for the outer-loop fixture pose selection and for the inner-loop robotic skill learning. We also use RL for learning a policy when gradually removing the fixture.
\begin{figure*}[hbt!]
\centering
    \centering
    \includegraphics[width=\linewidth]{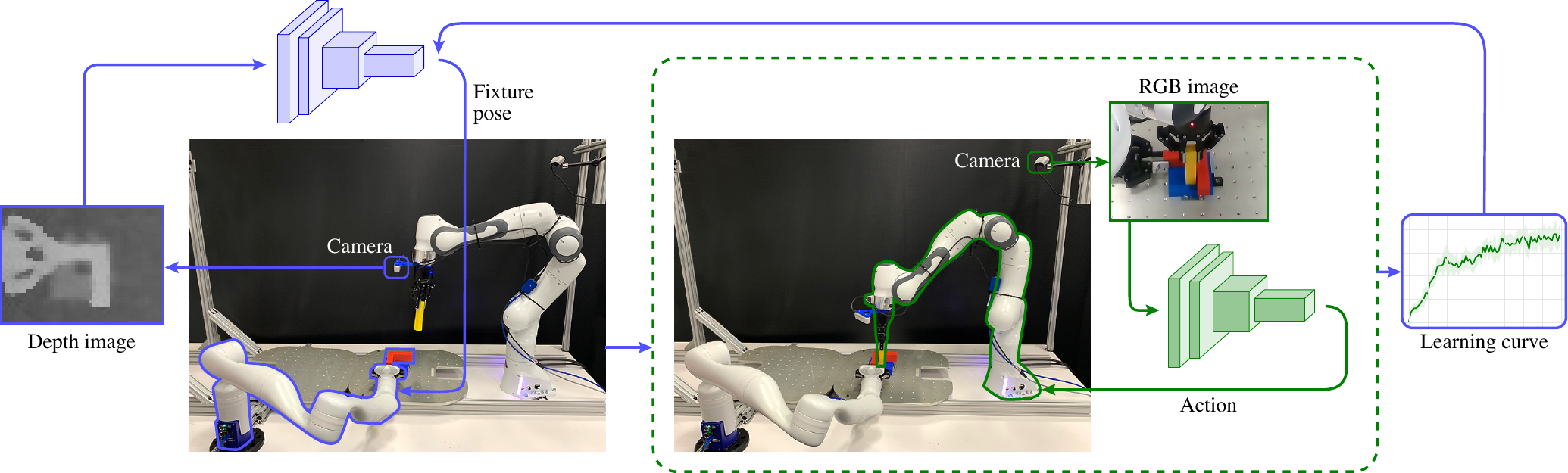}
    \caption{Overall Approach. The outer loop (blue arrows) takes a depth image of the manipulation scene and the learned policy proposes a fixture pose. After the fixture has been placed, the inner loop (in green dashed box) trains an RL policy that maps RGB images to end-effector motions to complete the manipulation task. The cumulative reward is sent to the outer loop to optimize the fixture placement. The goal of the outer loop is to maximize the learning rate and performance of the inner loop.}
    \label{fig::model}
\end{figure*}
In the following, $s \in \mathcal{S}$ denotes the state, which in our case is either a depth or an RGB image. $a \in \mathcal{A}$ denotes the action, which corresponds to the fixture pose in our outer-loop process or the end-effector motion in the inner-loop process. At each time step $t$, our robot perceives the state $s_t$ and chooses an action $a_t$. It receives a reward $R_t$ and moves to a new state $s_{t+1}$. The goal is to learn policies that maximize the cumulative reward.

\subsection{Fixture Pose Selection}
 The pose of a fixture is parameterized the vector $a^{f}$. 
 The robot perceives the manipulation scene through a top-view depth camera. 
 $s^f_0$ is the initial image and provides contextual information about the scene. The robot then selects a pose $a^{f}$ of the fixture and places the fixture in the workspace. The robot then starts the inner-loop learning process. After a fixed number of inner-loop episodes, the outer-loop receives a reward $R^{f}$ representing the inner-loop learning performance given the fixture pose $a^{f}$. To maximise $R^{f}$, the outer-loop needs to select the fixture pose such that the inner-loop masters the manipulation skill.
 


Because the contextual information in our problem is the initial depth image, we use a convolutional neural network model to capture the contextual features. Similar to Q-learning, we approximate the reward distribution given the contextual information. Let $Q(s^f_0,a^f)$ be the expected reward the robot receives after it sees the contextual information $s^f_0$ and takes the action $a^f$. We use a deep convolutional network denoted as $Q_w(s^f_0,a^f)$ to approximate $Q(s^f_0,a^f)$. We aim to train our model by minimizing the following loss:
 \begin{equation}
     \mathcal{L}^f = \E_{(s^f_0,a^f)} [Q_w(s^f_0,a^f)-Q(s^f_0,a^f)]
 \end{equation}
 
 At test time, given a depth image $s^f$, the optimal policy is $\pi(s^f)=\argmax_{a^f}Q_w(s^f,a^f)$. Similar to ~\citet{kalashnikov2018scalable}, we adopt the cross-entropy method (CEM)~\cite{Rubinstein2004TheCM} to perform the optimization with respect to action $a^f$. We first sample a batch of $M$ points in action space $\mathcal{A}$ and fit a Gaussian distribution based on the top $N$ samples. Given the fitted Gaussian, we repeated this sampling process six times. 

At training time, we aim to maximize the rewards after the robot perceives the contextual information. We propose a revised algorithm called \emph{Smoothed Zooming Algorithm} shown in Alg.\ref{alg:zooming}. We first cover the continuous action space with a countable number of small covering balls with the radius of $h$. We adopt the commonly used UCB1~\cite{auer2002finite} rule to select one covering ball. The fixture pose itself is uniformly sampled within the corresponding covering ball. In this way, we are able to smooth the discontinuous reward function~\cite{krishnamurthy2019contextual}. Each covering ball maintains the average reward $\bar{\mu}_t$ and the number of sampled points $n_t$ sampled from itself. As the number of sampled points increases, the radius of the ball $r_t$ shrinks and may lead to new exposure areas in the action space. We initialize new covering balls to cover these new exposure areas to ensure that these balls always cover the continuous action space. We repeat updating the center and radius of these covering balls in each round.
 

 
\begin{algorithm}
Initial radius of covering ball h $>$ 0, number of rounds $T^f$, Active Arm Set $S$, Arm $x \in \mathcal{X}$ continuous action space.\\
S = Set$\{\ceil[\big]{\frac{x}{h}}\}$, Activate these arms lying on the discrete mesh, $\bar\mu_0(x)_{x\in S}=0$
\begin{algorithmic}[1]
\For{$t$ in each round}
\State {$\bar{x}=\argmax_{x \in S}[\bar{\mu}_{t-1}(x)+2r_{t-1}(x)]$} \Comment{Selection}
    \State{$a_t \sim$ Uniform$(B_{t-1}(\bar{x}))$
    \State$n'_t(x)=\sum_{i=1,t}\mathbf{1}\{a_i \in B_{t-1}(x)\}$ 
    \State$r'_t(x)=\sqrt{\frac{2\log{T^f}}{n'_t(x)+1}}$
    \State$B_t(x)=\{y\in X: \mathcal{D}(x,y)\le r'_t(x)\}$ \Comment{Update Confidence Ball} 
    \State$n_t(x)=\sum_{i=1,t}\mathbf{1}\{a_i \in B_{t}(x)\}$
    \State$r_t(x)=\sqrt{\frac{2\log{T^f}}{n_t(x)+1}}$
    \State$\bar{\mu}_{t}(x)=\sum_{i=1,t}R_i(x)\mathbf{1}\{a_i \in B_t(x)\}/n_t(x)$}
\State{Activation: If some arm is not covered, pick any such arm and activate it.}
\EndFor
\end{algorithmic}
\caption{Smoothed Zooming Algorithm}\label{alg:zooming}
\end{algorithm}


\subsection{Robotic Skill Learning with Fixtures}~\label{sec:wf}
We formulate the inner learning loop for robot manipulation skills as an episodic RL problem. We denote $T$ as the maximum number of steps within each episode and $\gamma<1$ as the discount factor. At each time step $t \le T$, the robot receives an RGB image $s_t$ of the manipulation scene and performs an action $a_t$ at the end-effector. The robot moves to a new state $s_{t+1}$ and receives a reward $R_{t}$ from the environment. The episode ends either when the task is finished successfully or it reaches the maximum time step $T$. In each individual task, the robot learns a policy $\pi(s_t)$ and aims to maximize the cumulative reward. 
We use a variant of the A3C algorithm~\cite{mnih2016asynchronous} to learn the optimal policy in simulation. We use operational space control~\cite{khatib1987unified} for executing the devised end-effector motion and to decouple the constrained motion induced by the fixture and active force control in the manipulation task. 

\subsection{Robotic Skill Learning while Removing Fixtures}
When the robot learns a policy $\pi^*$ to finish the task successfully with the help of the fixture, we also want the robot to maintain this skill after the fixture is removed. To achieve this goal, we replace the hard constraints produced by the physical fixtures with soft constraints produced by a virtual potential field. Note that in the outer-loop stage, the robot perceives the manipulation scene based on a top-view depth camera. Given the depth image from that camera, we are able to infer the physical geometry of the fixture. Then we make the potential field to maintain the same geometry as the fixtures. 
We start to gradually remove the fixture away from the manipulation scene horizontally. Meanwhile, in our inner-loop process, we continue to train the policy starting from $\pi^*$. So we can transfer the manipulation skills with physical fixtures to skills with virtual potential fields automatically. For the inner-loop process, we maintain the same experiment setting. The removal speed is a hyper-parameter and may vary from different tasks. 

\section{Experiments}
We execute a series of experiments to test the proposed learning framework both in simulation and the real world. We investigate the following questions: 1) Do fixtures speed up manipulation skill learning? 2) Is the fixture pose selection policy reasonable? 3) Can the robot retain the learned manipulation skill after gradually removing the fixture? 4) How robust is the framework to task variations and perturbations? 

\begin{figure*}[htb!]
    \centering
    \includegraphics[width=1\linewidth]{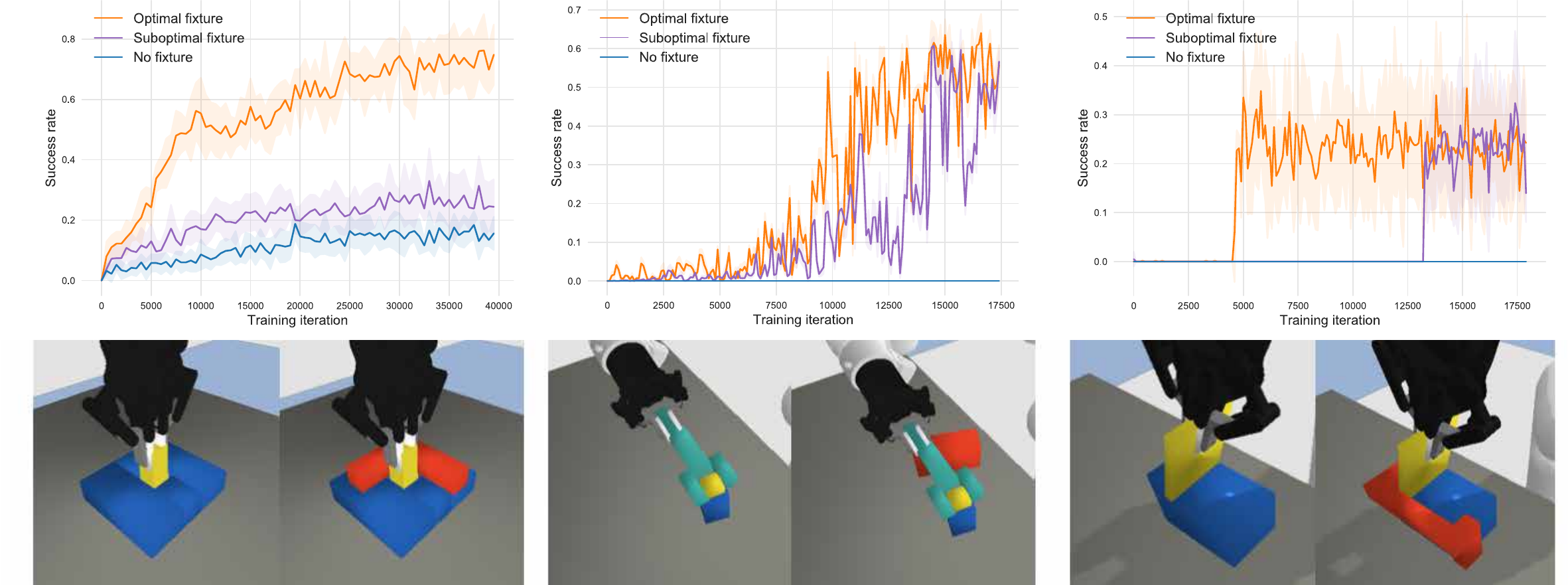}
    \caption{Learning curves for the three different tasks, along with visualizations of their simulation environments, both with and without the (red) fixtures. Success for peg insertion (left) and shallow-depth insertion (right) means completing the task before the end of the episode. 
    The suboptimal fixture pose is set to be one cm away from the optimal pose. 
    }
    \label{fig:learningCurves}
    \vspace{-0.3cm}
\end{figure*}

\subsection{Robotic Experiment Setup}
We use a 7-DoF Franka Panda robot arm with a two-fingered Robotiq 2F-85 gripper to perform the inner-loop manipulation tasks both in simulation and in the real world. We use PyBullet~\cite{coumans2019} as the simulation environment. An RGB-D camera is statically mounted in the environment to capture RGB images downsampled to $120 \times 160$ for the inner-loop manipulation tasks. 
The camera parameters are the same in simulation and the real world.

For the outer-loop fixture pose selection, we directly control the fixture pose in simulation and fix it at the given pose. In the real-world experiment, we use a 7-DoF Kinova Gen3 arm with a Robotiq 2F-85 gripper to position and firmly hold the fixture. In both simulation and the real world, we use a wrist-mounted RGB-D camera to obtain depth images downsampled to $60 \times 80$ for the outer-loop task.

We design three different manipulation tasks as shown in Fig.~\ref{fig:removal}: inserting a peg into a hole (denoted as \textbf{Insertion}), tightening a bolt using a wrench (\textbf{Wrench}) and inserting a thin cuboid into a shallow slot~\cite{kim2019shallow} (\textbf{SD Insertion}). 

The actions are described in the robot base frame. In the peg-insertion task, we set the inner-loop action $a\in \mathcal{R}^3$ to be the relative translation motion of the end-effector. The fixture pose is also a three dimensional vector $a^f=(x,y,\theta)$ representing the fixture's 2D position and orientation in the plane perpendicular to the axis of the hole. In the wrench manipulation task, the inner-loop action is a 6D relative motion vector $(dx,dy,dz,d\alpha,d\beta,d\gamma)$, and the outer-loop action is the 3D position $(x,y,z)$ of the fixture. 
For the shallow-depth insertion~\cite{kim2019shallow} task, the inner-loop action is the relative 2D translation and orientation of the end-effector in the plane perpendicular to the axis along the short width of the hole.

The fixture in the peg-insertion task helps to prevent the peg from slipping past the hole. In the wrench manipulation task, the fixture provides support for the wrench to stay horizontal and apply a continuous torque on the bolt. In the shallow-depth insertion~\cite{kim2019shallow} task, the fixture restricts the translation and rotation motion of the cuboid into the slot when the gripper releases and pulls away from the cuboid.  

\subsection{Evaluation Metrics and Reward Design}
The peg-insertion task is considered to be successful if the peg is completely inserted into the hole. The wrench manipulation task is successful if the bolt is rotated by 30 degrees. The shallow-depth insertion task successful if the cuboid is horizontal and inside the shallow slot. We report the success rates for the peg-insertion and shallow-depth insertion tasks, and the turning rate of the bolt for the wrench manipulation task.

We have the following task-specific reward functions:
\begin{align}
&R(\textbf{Insertion})=
\left\{
\begin{array}{lll}
      &  5.0 &  \text{if success}\\
      &- 1.0 & \text{if dist(peg, hole)} > \text{4cm}\\
      &  0.0 & \text{otherwise} \\
\end{array} 
\right.\\
&R(\textbf{Wrench}) = 
\left\{
\begin{array}{lll}
      &5.0 & \text{if success} \\
      &- 1.0 & \text{if dist(wrench head, bolt)} > \text{4cm}\\
      &\eta &\text{otherwise:} \\
\end{array} 
\right.\\
&R(\textbf{SDInsertion})= 
\left\{
\begin{array}{lll}
      &5.0 \hspace{0.3cm} \text{if success} & \\
      &-1.0 \hspace{0.2cm}  \text{if dist}(\text{battery},\text{phone}) > \text{4cm}&\\
      &\text{otherwise:} &\\
 &0.1\times \exp{(-\text{dist}(\text{battery},\text{phone})}+\zeta & \\
\end{array} 
\right.
\end{align}
where $\eta$ and $\zeta$ represent the rotation degree of the bolt and battery.

\subsection{Task Performance Evaluation with and without Fixtures}
In this simulation experiment, we evaluate whether optimally-positioned physical fixtures speed up manipulation skill learning. During training, we randomize the configuration of the absolute object position and the arm's initial position at the beginning of each episode. 

We illustrate the training curves in Fig.~\ref{fig:learningCurves}. We train the polices with 40k, 17.5k, and 17.5k steps for the insertion, wrench, and shallow-insertion tasks, respectively, and save the trained models at 500, 100, and 100 steps, respectively. Then, we evaluate the trained models for 5k steps on three versions of the task: one with the fixture at the optimal pose, one with the fixture at a suboptimal pose, and one with no fixture at all. From Fig.~\ref{fig:learningCurves}, we observe that a proper fixture significantly improves both task performance and learning speed for all three tasks. For the insertion task, the agent achieves a nearly 80\% success rate with the optimal fixture pose but can only achieve 20\% without a fixture. In order to achieve a 20\% success rate, the agent needs to run 40k steps without a fixture but only 5k steps with an optimal fixture pose. Wrench manipulation and shallow-depth insertion are challenging manipulation tasks. We observes that the robots can only achieve reasonable progress with the help of fixtures.

\subsection{Fixture Pose Selection and Visualization}
\begin{figure}[ht!]
    \centering
    \includegraphics[width=1\linewidth]{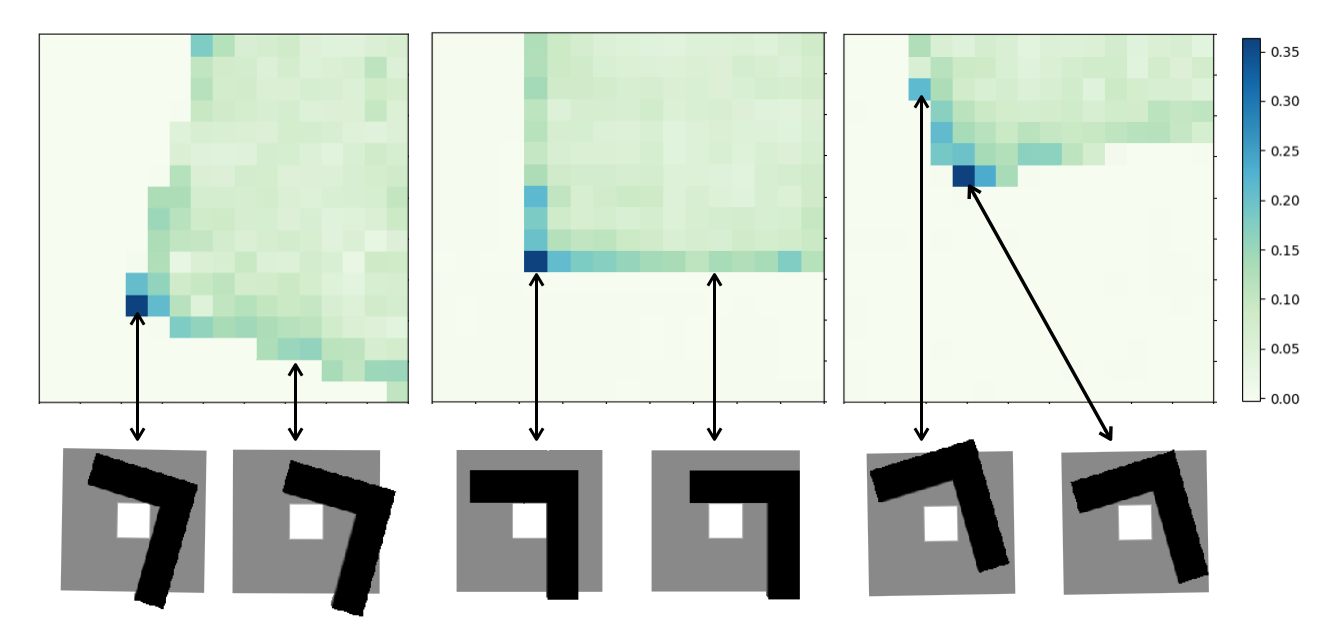}
    \caption{Visualization of the Q-function for the policy of the outer loop. Each pixel in the three images corresponds to a $x,y$ fixture position with an orientation of either $-15$, $0$ or $15$ degrees. The color corresponds to the predicted reward where darker means higher.}
    \label{fig:Qmap}
\end{figure}

We visualize the learned Q map for the insertion task in Fig.~\ref{fig:Qmap}. Note that the outer-loop action is parameterized by a three-dimensional vector $(x,y,\theta)$ representing horizontal translation and rotation in the plane perpendicular to the hole. From left to right, each image represents the Q-function where the fixture is set to be $(u,v,-15^{\circ})$, $(u,v,0^{\circ})$ and $(u,v,15^{\circ})$. Each pixel $u,v$ corresponds to an $x,y$ position of the fixture and is colored according to the predicted reward - the darker, the higher. We also visualize for some pixels what the corresponding fixture pose is, thereby giving an intuitive understanding of the reward.

The three images indicate that our Q function learns a reasonable policy for the fixture pose selection. As expected, the highest Q value corresponds to the optimal pose where the fixture is lined up with the hole. The continuous gradual change up to the optimal pose allows CEM~\cite{Rubinstein2004TheCM} to find it.

\subsection{Task Performance Evaluation with Fixture Removal}\label{sec:Fremoval}
\begin{figure}[t!]
    \centering
    \includegraphics[width=1\linewidth]{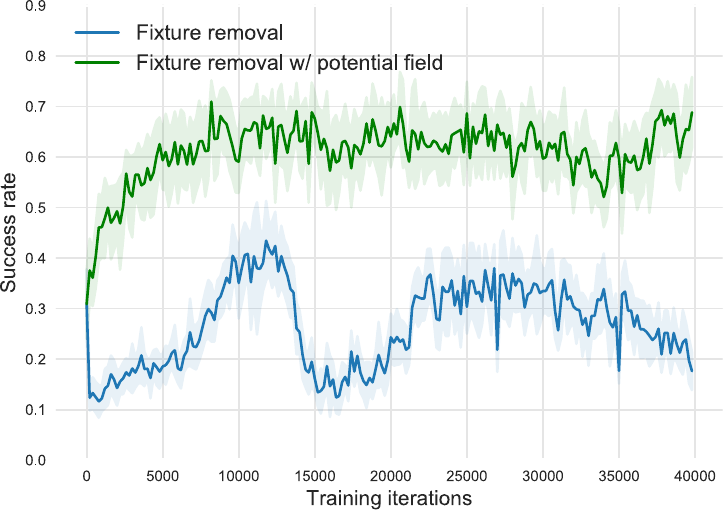}
    \caption{Performance on peg-in-hole task with fixture removal.} 
    \label{fig:removal}
    \vspace{-0.3cm}
\end{figure}
After training the agent on the peg-in-hole task with a fixture, we also want to maintain its task performance while gradually removing the fixture. In the baseline method, we continue to train the agent while gradually removing the fixture away from the hole by 1cm every 2k steps. We compare this to a method where we similarly remove the physical fixture, but also keep a soft constraint at the original location of the physical fixture in the form of a repulsive potential field (see Fig.~\ref{fig:removal}). We find that with the baseline method, performance initially drops to 30\%, but drops further to around 10\% as the fixture moves farther away. With the potential field, the performance still drops to 30\% initially, but climbs back up to 70\% when the fixture is fully removed as it learns to handle the absence of the physical fixture. 

\subsection{Demonstrating Generalization Ability}
\begin{figure}[t]
\begin{minipage}{\linewidth}
     \centering
      \includegraphics[width=\linewidth]{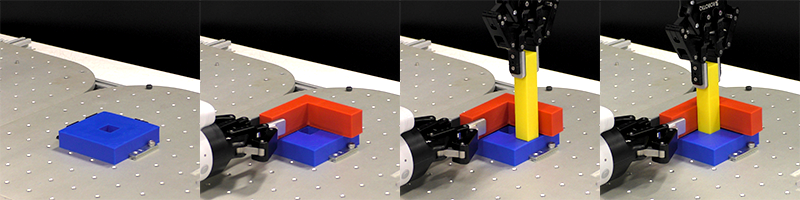}
      \caption{\label{fig:real_peg}Insert a round peg in round hole}
     \end{minipage}\\
       \begin{minipage}{\linewidth}
     \centering
      \includegraphics[width=\linewidth]{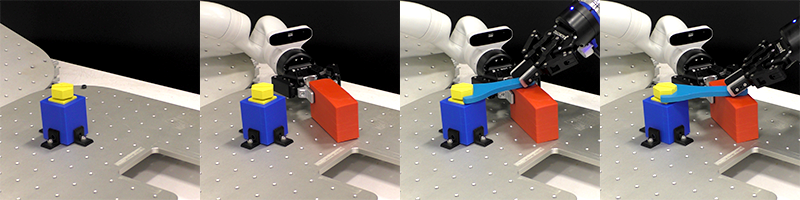}
       \caption{\label{fig:real_wrench}Rotate a bolt using a wrench}
     \end{minipage}\\
            \begin{minipage}{\linewidth}
     \centering
      \includegraphics[width=\linewidth]{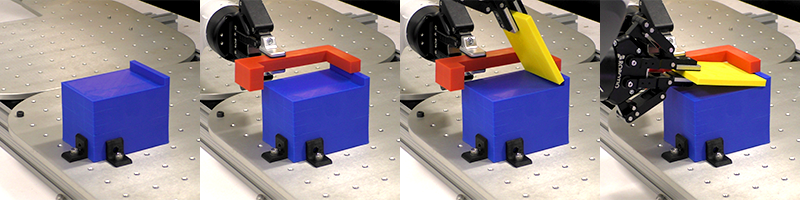}
       \caption{\label{fig:real_shallow}Rotate and insert a thin, flat cuboid into a shallow hole}
     \end{minipage}\\
     \label{fig:real}
\end{figure}

We directly test the simulation-trained policy in the real world. We utilize RGB images to segment objects and use these segmentation masks to improve the quality of depth images. To reduce the domain gap between simulation and the real world, we utilize domain randomization~\cite{tobin2017domain} approaches and randomize object positions, textures, and lighting conditions. The real world experiment setting is shown in Fig.~\ref{fig::model}. Snapshots of task performance are shown in Figs.~\ref{fig:real_peg},\ref{fig:real_wrench} and \ref{fig:real_shallow}. For each task, we compare the robustness of different methods by changing the positions of objects in different trials. The results are presented in Fig.~\ref{fig:result}. The models trained without fixtures fail in all tests, reflecting the difficulty of these manipulation tasks. The models trained with fixtures successfully finish almost every trial for each task, if not all. We also compare the models trained on fixture removal (Sec.~\ref{sec:Fremoval}) for the peg-insertion task. The baseline fails to insert the peg into the hole, while our model trained with the potential field is able to maintain the skill. We also test the generalization of our proposed method over different peg geometry, as shown in the accompanying video.

\begin{figure}[ht!]
  \centering 
 \begin{tabular}{c | c c c c}
\hline
\hline 
Method  & \textbf{Insertion} & \textbf{Wrench} & \textbf{SD Insertion}\\ 
\hline
w/o Fixture &0/5& 0/5 & 0/5 \\
w/ Fixture &5/5& 5/5 & 4/5  \\
-Fixture & 1/5 &- & - \\
-Fixture+PF & 5/5 & - & -  \\
\hline 
\end{tabular}
\caption{Comparison of success rates for different tasks. w/o Fixture refers to policies learned completely without fixtures. w/ Fixture refers to our proposed method utilizing fixtures. -Fixtures refers to the fixture removal baseline in Sec.~\ref{sec:Fremoval}. -Fixtures+PF refers to our proposed fixture removal method with potential fields method in Sec.~\ref{sec:Fremoval}.}
\label{fig:result}
\end{figure}

\label{sec: exp}

\section{Conclusion}
Automatically learning robotic manipulation skills is a challenging problem due to high-dimensional state and action spaces, noisy sensors, and inaccurate motor control. We proposed a learning framework which enables a robot to autonomously change its environment and discover how to expedite manipulation skill learning. We provide the robot with fixtures that it can freely place within the environment to generate motion constraints that limit the outcome of robot actions. The fixtures funnel uncertainty from perception and motor control and {\em scaffold\/} manipulation skill learning. We show in simulation and in the real world that our framework dramatically speeds up robotic skills on three tasks: peg insertion, wrench manipulation, and shallow-depth insertion~\cite{kim2019shallow}. 

\label{sec:conclusion}

{\small
\bibliographystyle{IEEEtranN}
\bibliography{references}

\begin{thebibliography}{25}
\providecommand{\natexlab}[1]{#1}
\providecommand{\url}[1]{#1}
\csname url@samestyle\endcsname
\providecommand{\newblock}{\relax}
\providecommand{\bibinfo}[2]{#2}
\providecommand{\BIBentrySTDinterwordspacing}{\spaceskip=0pt\relax}
\providecommand{\BIBentryALTinterwordstretchfactor}{4}
\providecommand{\BIBentryALTinterwordspacing}{\spaceskip=\fontdimen2\font plus
\BIBentryALTinterwordstretchfactor\fontdimen3\font minus
  \fontdimen4\font\relax}
\providecommand{\BIBforeignlanguage}[2]{{%
\expandafter\ifx\csname l@#1\endcsname\relax
\typeout{** WARNING: IEEEtranN.bst: No hyphenation pattern has been}%
\typeout{** loaded for the language `#1'. Using the pattern for}%
\typeout{** the default language instead.}%
\else
\language=\csname l@#1\endcsname
\fi
#2}}
\providecommand{\BIBdecl}{\relax}
\BIBdecl

\bibitem[Deimel et~al.(2016)Deimel, Eppner, {\'A}lvarez-Ruiz, Maertens, and
  Brock]{deimel2016exploitation}
R.~Deimel, C.~Eppner, J.~{\'A}lvarez-Ruiz, M.~Maertens, and O.~Brock,
  ``Exploitation of environmental constraints in human and robotic grasping,''
  in \emph{Robotics Research}.\hskip 1em plus 0.5em minus 0.4em\relax Springer,
  2016, pp. 393--409.

\bibitem[Kazemi et~al.()Kazemi, Valois, Bagnell, and Pollard]{kazemi2012robust}
M.~Kazemi, J.-S. Valois, J.~A. Bagnell, and N.~Pollard, ``Robust object
  grasping using force compliant motion primitives.''

\bibitem[Righetti et~al.(2014)Righetti, Kalakrishnan, Pastor, Binney, Kelly,
  Voorhies, Sukhatme, and Schaal]{righetti2014autonomous}
L.~Righetti, M.~Kalakrishnan, P.~Pastor, J.~Binney, J.~Kelly, R.~C. Voorhies,
  G.~S. Sukhatme, and S.~Schaal, ``An autonomous manipulation system based on
  force control and optimization,'' \emph{Autonomous Robots}, vol.~36, no. 1-2,
  pp. 11--30, 2014.

\bibitem[Hudson et~al.(2012)Hudson, Howard, Ma, Jain, Bajracharya, Myint, Kuo,
  Matthies, Backes, Hebert, et~al.]{hudson2012end}
N.~Hudson, T.~Howard, J.~Ma, A.~Jain, M.~Bajracharya, S.~Myint, C.~Kuo,
  L.~Matthies, P.~Backes, P.~Hebert \emph{et~al.}, ``End-to-end dexterous
  manipulation with deliberate interactive estimation,'' in \emph{2012 IEEE
  International Conference on Robotics and Automation}.\hskip 1em plus 0.5em
  minus 0.4em\relax IEEE, 2012, pp. 2371--2378.

\bibitem[Toussaint et~al.(2014)Toussaint, Ratliff, Bohg, Righetti, Englert, and
  Schaal]{toussaint2014dual}
M.~Toussaint, N.~Ratliff, J.~Bohg, L.~Righetti, P.~Englert, and S.~Schaal,
  ``Dual execution of optimized contact interaction trajectories,'' in
  \emph{2014 IEEE/RSJ International Conference on Intelligent Robots and
  Systems}.\hskip 1em plus 0.5em minus 0.4em\relax IEEE, 2014, pp. 47--54.

\bibitem[Chavan-Dafle and Rodriguez()]{chavan2015prehensile}
N.~Chavan-Dafle and A.~Rodriguez, ``Prehensile pushing: In-hand manipulation
  with push-primitives,'' in \emph{2015 IEEE/RSJ International Conference on
  Intelligent Robots and Systems (IROS)}.\hskip 1em plus 0.5em minus
  0.4em\relax IEEE, pp. 6215--6222.

\bibitem[Asada and By(1985)]{asada1985kinematic}
H.~Asada and A.~By, ``Kinematic analysis of workpart fixturing for flexible
  assembly with automatically reconfigurable fixtures,'' \emph{IEEE Journal on
  Robotics and Automation}, vol.~1, no.~2, pp. 86--94, 1985.

\bibitem[Chou et~al.(1989)Chou, Chandru, and Barash]{chou1989mathematical}
Y.-C. Chou, V.~Chandru, and M.~M. Barash, ``A mathematical approach to
  automatic configuration of machining fixtures: analysis and synthesis,''
  \emph{Journal of Engineering for Industry}, vol. 111, no.~4, pp. 299--306,
  1989.

\bibitem[Kim and Seo(2019)]{kim2019shallow}
C.~H. Kim and J.~Seo, ``Shallow-depth insertion: Peg in shallow hole through
  robotic in-hand manipulation,'' \emph{IEEE Robotics and Automation Letters},
  vol.~4, no.~2, pp. 383--390, 2019.

\bibitem[Inoue et~al.(2017)Inoue, De~Magistris, Munawar, Yokoya, and
  Tachibana]{inoue2017deep}
T.~Inoue, G.~De~Magistris, A.~Munawar, T.~Yokoya, and R.~Tachibana, ``Deep
  reinforcement learning for high precision assembly tasks,'' in \emph{2017
  IEEE/RSJ International Conference on Intelligent Robots and Systems
  (IROS)}.\hskip 1em plus 0.5em minus 0.4em\relax IEEE, 2017, pp. 819--825.

\bibitem[Thomas et~al.(2018)Thomas, Chien, Tamar, Ojea, and
  Abbeel]{thomas2018learning}
G.~Thomas, M.~Chien, A.~Tamar, J.~A. Ojea, and P.~Abbeel, ``Learning robotic
  assembly from cad,'' in \emph{2018 IEEE International Conference on Robotics
  and Automation (ICRA)}.\hskip 1em plus 0.5em minus 0.4em\relax IEEE, 2018,
  pp. 1--9.

\bibitem[Luo et~al.(2019)Luo, Solowjow, Wen, Ojea, Agogino, Tamar, and
  Abbeel]{luo2019reinforcement}
J.~Luo, E.~Solowjow, C.~Wen, J.~A. Ojea, A.~M. Agogino, A.~Tamar, and
  P.~Abbeel, ``Reinforcement learning on variable impedance controller for
  high-precision robotic assembly,'' \emph{arXiv preprint arXiv:1903.01066},
  2019.

\bibitem[Lee et~al.(2019)Lee, Zhu, Srinivasan, Shah, Savarese, Fei-Fei, Garg,
  and Bohg]{lee2019making}
M.~A. Lee, Y.~Zhu, K.~Srinivasan, P.~Shah, S.~Savarese, L.~Fei-Fei, A.~Garg,
  and J.~Bohg, ``Making sense of vision and touch: Self-supervised learning of
  multimodal representations for contact-rich tasks,'' in \emph{2019
  International Conference on Robotics and Automation (ICRA)}.\hskip 1em plus
  0.5em minus 0.4em\relax IEEE, 2019, pp. 8943--8950.

\bibitem[Zhu et~al.(2015)Zhu, Zhao, and Chun~Zhu]{zhu2015understanding}
Y.~Zhu, Y.~Zhao, and S.~Chun~Zhu, ``Understanding tools: Task-oriented object
  modeling, learning and recognition,'' in \emph{Proceedings of the IEEE
  Conference on Computer Vision and Pattern Recognition}, 2015, pp. 2855--2864.

\bibitem[Fang et~al.(2018)Fang, Zhu, Garg, Kurenkov, Mehta, Fei-Fei, and
  Savarese]{fang2018learning}
K.~Fang, Y.~Zhu, A.~Garg, A.~Kurenkov, V.~Mehta, L.~Fei-Fei, and S.~Savarese,
  ``Learning task-oriented grasping for tool manipulation from simulated
  self-supervision,'' \emph{arXiv preprint arXiv:1806.09266}, 2018.

\bibitem[Kleinberg et~al.(2013)Kleinberg, Slivkins, and
  Upfal]{kleinberg2013bandits}
R.~Kleinberg, A.~Slivkins, and E.~Upfal, ``Bandits and experts in metric
  spaces,'' \emph{arXiv preprint arXiv:1312.1277}, 2013.

\bibitem[Krishnamurthy et~al.(2019)Krishnamurthy, Langford, Slivkins, and
  Zhang]{krishnamurthy2019contextual}
A.~Krishnamurthy, J.~Langford, A.~Slivkins, and C.~Zhang, ``Contextual bandits
  with continuous actions: Smoothing, zooming, and adapting,'' \emph{arXiv
  preprint arXiv:1902.01520}, 2019.

\bibitem[Dudik et~al.(2011)Dudik, Hsu, Kale, Karampatziakis, Langford, Reyzin,
  and Zhang]{dudik2011efficient}
M.~Dudik, D.~Hsu, S.~Kale, N.~Karampatziakis, J.~Langford, L.~Reyzin, and
  T.~Zhang, ``Efficient optimal learning for contextual bandits,'' \emph{arXiv
  preprint arXiv:1106.2369}, 2011.

\bibitem[Kalashnikov et~al.(2018)Kalashnikov, Irpan, Pastor, Ibarz, Herzog,
  Jang, Quillen, Holly, Kalakrishnan, Vanhoucke,
  et~al.]{kalashnikov2018scalable}
D.~Kalashnikov, A.~Irpan, P.~Pastor, J.~Ibarz, A.~Herzog, E.~Jang, D.~Quillen,
  E.~Holly, M.~Kalakrishnan, V.~Vanhoucke \emph{et~al.}, ``Scalable deep
  reinforcement learning for vision-based robotic manipulation,'' in
  \emph{Conference on Robot Learning}, 2018, pp. 651--673.

\bibitem[Rubinstein and Kroese(2004)]{Rubinstein2004TheCM}
R.~Y. Rubinstein and D.~P. Kroese, ``The cross-entropy method,'' in
  \emph{Information Science and Statistics}, 2004.

\bibitem[Auer et~al.(2002)Auer, Cesa-Bianchi, and Fischer]{auer2002finite}
P.~Auer, N.~Cesa-Bianchi, and P.~Fischer, ``Finite-time analysis of the
  multiarmed bandit problem,'' \emph{Machine learning}, vol.~47, no. 2-3, pp.
  235--256, 2002.

\bibitem[Mnih et~al.(2016)Mnih, Badia, Mirza, Graves, Lillicrap, Harley,
  Silver, and Kavukcuoglu]{mnih2016asynchronous}
V.~Mnih, A.~P. Badia, M.~Mirza, A.~Graves, T.~Lillicrap, T.~Harley, D.~Silver,
  and K.~Kavukcuoglu, ``Asynchronous methods for deep reinforcement learning,''
  in \emph{International conference on machine learning}, 2016, pp. 1928--1937.

\bibitem[Khatib(1987)]{khatib1987unified}
O.~Khatib, ``A unified approach for motion and force control of robot
  manipulators: The operational space formulation,'' \emph{IEEE Journal on
  Robotics and Automation}, vol.~3, no.~1, pp. 43--53, 1987.

\bibitem[Coumans and Bai(2016--2019)]{coumans2019}
E.~Coumans and Y.~Bai, ``Pybullet, a python module for physics simulation for
  games, robotics and machine learning,'' \url{http://pybullet.org},
  2016--2019.

\bibitem[Tobin et~al.(2017)Tobin, Fong, Ray, Schneider, Zaremba, and
  Abbeel]{tobin2017domain}
J.~Tobin, R.~Fong, A.~Ray, J.~Schneider, W.~Zaremba, and P.~Abbeel, ``Domain
  randomization for transferring deep neural networks from simulation to the
  real world,'' in \emph{2017 IEEE/RSJ International Conference on Intelligent
  Robots and Systems (IROS)}.\hskip 1em plus 0.5em minus 0.4em\relax IEEE,
  2017, pp. 23--30.

\end{thebibliography}
}

\end{document}